\documentclass{article}

\usepackage[preprint]{neurips_2025}

\usepackage[utf8]{inputenc} 
\usepackage[T1]{fontenc}    
\usepackage{hyperref}       
\usepackage{url}            
\usepackage{booktabs}       
\usepackage{amsfonts}       
\usepackage{nicefrac}       
\usepackage{microtype}      
\usepackage{xcolor}         
\newtheorem{definition}{Definition}[section]
\newtheorem{proposition}{Proposition}[section]
\usepackage{graphicx}
\usepackage{algorithm}
\usepackage{algorithmic}
\usepackage{multirow}
\usepackage{amsmath}

\title{Governing Equation Discovery from Data Based on Differential Invariants}

\author{%
  Lexiang Hu \\
  Peking University \\
  \texttt{hulx@stu.pku.edu.cn} \\
  \And
  Yikang Li \\
  Peking University \\
  \texttt{liyk18@pku.edu.cn}
  \AND
  Zhouchen Lin \\
  Peking University \\
  \texttt{zlin@pku.edu.cn} \\
}

\begin{document}

\maketitle

\begin{abstract}
  The explicit governing equation is one of the simplest and most intuitive forms for characterizing physical laws. However, directly discovering partial differential equations (PDEs) from data poses significant challenges, primarily in determining relevant terms from a vast search space. Symmetry, as a crucial prior knowledge in scientific fields, has been widely applied in tasks such as designing equivariant networks and guiding neural PDE solvers. In this paper, we propose a pipeline for governing equation discovery based on differential invariants, which can losslessly reduce the search space of existing equation discovery methods while strictly adhering to symmetry. Specifically, we compute the set of differential invariants corresponding to the infinitesimal generators of the symmetry group and select them as the relevant terms for equation discovery. Taking DI-SINDy (SINDy based on Differential Invariants) as an example, we demonstrate that its success rate and accuracy in PDE discovery surpass those of other symmetry-informed governing equation discovery methods across a series of PDEs.
\end{abstract}

\section{Introduction}
\label{sec:introduction}

Explicit equations, particularly partial differential equations (PDEs), play a significant role in scientific fields due to their concise and intuitive mathematical forms. Discovering governing equations directly from observational data has become an important topic, and its solutions may serve as AI assistants to human scientists in uncovering new physical laws. Although neural PDE solvers also aim for data-driven evolution prediction \citep{greydanus2019hamiltonian,bar2019learning,sanchez2020learning,li2020fourier,thuerey2021physics,brandstetter2022message,gupta2022towards,takamoto2022pdebench,takamoto2023learning,lippe2023pde,kapoor2023neural,cho2024parameterized,musekamp2024active}, their implicit learning approach, compared to explicit equation discovery, suffers from limitations such as lack of interpretability and weaker out-of-distribution (OOD) generalization. In this paper, we formalize the problem as discovering the governing PDE $F(x, u^{(n)})=0$ from trajectory data $u(x)$, where $x \in \mathbb{R}^p$ represents the independent variables, $u \in \mathbb{R}^q$ denotes the dependent variables, and $u^{(n)}$ signifies derivatives of $u$ with respect to $x$ up to order $n$.

Some previous works have made progress on the data-driven equation discovery problem. One category of search-based methods \citep{schmidt2009distilling,gaucel2014learning,petersen2019deep,cranmer2019learning,cranmer2020discovering,udrescu2020ai,la2021contemporary,mundhenk2021symbolic,sun2022symbolic,cranmer2023interpretable} explores the structure of equations interpretably, but their enormous search space incurs high computational costs. Another category of deep learning-based approaches \citep{brunton2016discovering,champion2019data,biggio2021neural,messenger2021weak,kamienny2022end} is generally more efficient and versatile, yet still requires pre-specifying key relevant terms of the equation skeleton. To address the limitations of these works, we need to leverage prior knowledge of scientific problems to constrain the form of equations---in other words, to narrow the search space of equations.

Symmetry is important prior knowledge in scientific problems, with each symmetry corresponding to a conserved quantity. Recently, some studies have attempted to discover symmetries from data for symmetry-dependent downstream tasks \citep{benton2020learning,dehmamy2021automatic,moskalev2022liegg,desai2022symmetry,yang2023generative,yang2023latent,ko2024learning,shaw2024symmetry}. Our goal is to leverage known symmetries to guide the discovery of governing equations. Although \citet{yang2024symmetry} achieve this by adding explicit symmetry constraints or implicit symmetry regularization terms, the governing equations they identify cannot strictly adhere to general symmetries, and the manually specified equation skeletons significantly affect accuracy.

\begin{figure}[h]
	\centering
	\includegraphics[width=\textwidth]{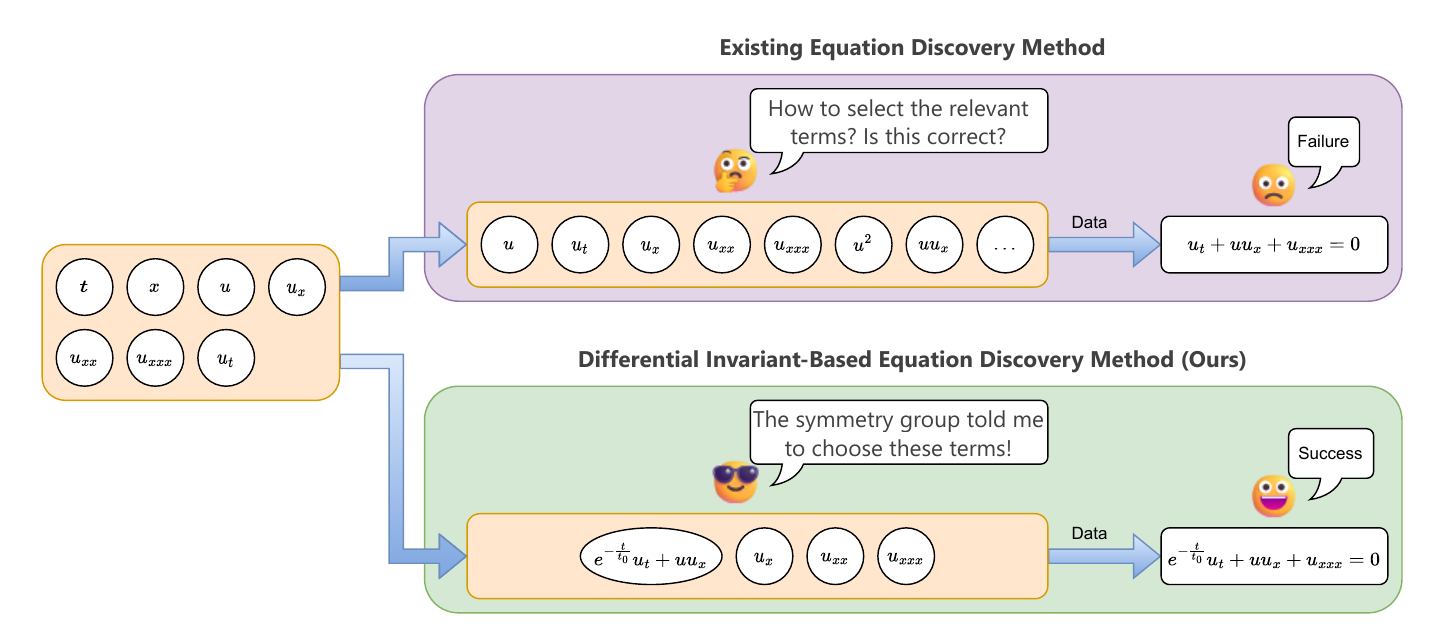}
	
	\caption{Comparison between the existing equation discovery method and our differential invariant-based equation discovery method for the nKdV equation $e^{-\frac{t}{t_0}} u_t + u u_x + u_{xxx} = 0$. The former struggles with selecting relevant terms, whereas our relevant terms are directly determined by the symmetry group.}
	\label{fig:introduction}
\end{figure}

In this paper, we implement symmetry-guided equation discovery based on differential invariants. Given the infinitesimal generators of a symmetry group, we can derive their prolongation forms and differential invariants. Then, we directly select these differential invariants as the relevant terms and plug them into any existing equation discovery method, such as SINDy \citep{brunton2016discovering}. The proposition cited in Section \ref{sec:method2} will demonstrate that this approach hard-embeds symmetry into the equation skeleton without sacrificing its expressive power. In other words, we ``losslessly'' compress the search space of equations. As shown in Figure \ref{fig:introduction}, for the relatively complex nKdV equation $e^{-\frac{t}{t_0}} u_t + u u_x + u_{xxx} = 0$, existing equation discovery methods struggle to identify the key relevant terms and construct the correct equation skeleton from a large search space of partial derivatives, whereas our method can accurately determine it by leveraging the information of the symmetry group.

In summary, our contributions are as follows: (1) we propose a method for equation discovery based on differential invariants, which is guided by symmetry groups in the selection of key relevant terms; (2) using the existing proposition, we substantiate that our method ensures the equation skeleton strictly adheres to symmetry without compromising its expressive power; (3) taking SINDy based on Differential Invariants (DI-SINDy) as an example, we demonstrate that our method can be plug-and-play with existing equation discovery approaches; (4) the experimental results on a series of PDEs show that our DI-SINDy achieves higher success rates and accuracy compared with baseline methods, while also exhibiting greater stability in long-term predictions.

\section{Related work}

\paragraph{Symmetry discovery.}

The application of symmetry in downstream tasks is based on the premise that we know it in advance; otherwise, we first need to discover the symmetry from the data. Some works discover symmetry based on Lie group and Lie algebra representations \citep{dehmamy2021automatic,moskalev2022liegg,desai2022symmetry,yang2023generative}, but they are limited to linear symmetries. Subsequent works attempt to find more complex nonlinear symmetries \citep{yang2023latent,ko2024learning,shaw2024symmetry}. They utilize the discovered symmetries to guide downstream tasks, achieving performance improvements, which validates the effectiveness of the results. The techniques in this paper can be combined with these symmetry discovery methods to address scenarios where symmetries are not known in advance.

\paragraph{Governing equation discovery.}

Automatically discovering governing equations from data is an important topic at the intersection of AI and science. One branch of methods relies on search algorithms and has achieved interpretable results. Deep Symbolic Regression (DSR) \citep{petersen2019deep} employs a novel risk-seeking policy gradient to train a recurrent neural network, which emits a distribution over tractable mathematical expressions. \citet{mundhenk2021symbolic} utilize neural-guided search to generate starting populations for a random restart genetic programming component, aiming to solve symbolic regression and other symbolic optimization problems. Symbolic Physics Learner (SPL) \citep{sun2022symbolic} machine leverages a Monte Carlo tree search (MCTS) agent to construct optimal expression trees, which interpret mathematical operations and variables. PySR \citep{cranmer2023interpretable} adopts a multi-population evolutionary algorithm and a unique evolve-simplify-optimize loop to accelerate the discovery of symbolic models. However, a limitation of such methods is their low computational efficiency when the search space is large.

Another branch of methods leverages deep learning to improve the efficiency of equation discovery. SINDy \citep{brunton2016discovering} employs sparse regression to identify equation forms that are both accurate and concise. Building upon SINDy, \citet{champion2019data} further utilize a deep autoencoder network to transform coordinates into a reduced space where the dynamics can be sparsely represented. Weak SINDy \citep{messenger2021weak} replaces pointwise derivative approximations with linear transformations and variance reduction techniques to enhance the robustness of SINDy against noise. NeSymReS \citep{biggio2021neural} pre-trains a Transformer to predict from an unbounded set of equations. These methods still require assumptions about key relevant terms of the equation skeleton and fail to incorporate scientific prior knowledge to narrow the search space for equations.

\paragraph{Applications of symmetry.}

Symmetry plays an important role in both traditional mathematical physics problems and the field of deep learning. We summarize related works in Appendix \ref{sec:application}.

\section{Preliminary}
\label{sec:preliminary}

Before introducing the method, we will first briefly present some preliminary knowledge concerning partial differential equations and their Lie point symmetries. For more details, please refer to the textbook \citep{olver1993applications}. We provide concrete examples in Appendix \ref{sec:example} to help readers better understand these concepts intuitively.

\paragraph{Partial differential equations.}

Let the independent variable $x \in X = \mathbb{R}^p$ and the dependent variable $u \in U = \mathbb{R}^q$. We denote the $k$-th order derivative of $u$ with respect to $x$ as $u_J^\alpha = \frac{\partial^k u^\alpha}{\partial x^{j_1} \partial x^{j_2} \dots \partial x^{j_k}} \in U_k$, where $\alpha \in \{1, \dots, q\}$, $J=(j_1, \dots, j_k)$, and $j_i \in \{1, \dots, p\}$. Furthermore, all derivatives of $u$ with respect to $x$ up to order $n$ are denoted as $u^{(n)} \in U^{(n)} = U \times U_1 \times \dots \times U_n$. Based on the above concepts, we can define a system of $n$-th order partial differential equations as $F(x, u^{(n)}) = 0$, where $F: X \times U^{(n)} \rightarrow \mathbb{R}^l$. Its solution is given by a smooth function $f: X \rightarrow U$.

\paragraph{Lie point symmetries.}

The solution to the system of partial differential equations $F(x, u^{(n)}) = 0$ can also be represented by the graph $\Gamma_f = \{(x, f(x)): x \in X\}$ of the function $f: X \rightarrow U$. Let the Lie group $G$ act on $X \times U$. We say that $G$ is a symmetry group of $F(x, u^{(n)}) = 0$ if, for any solution $f$ with its graph $\Gamma_f$ and any group element $g \in G$, $g \cdot \Gamma_f = \{(\tilde{x}, \tilde{u}) = g \cdot (x, u): (x, u) \in \Gamma_f\}$ is the graph $\Gamma_{\tilde{f}}$ of another solution $\tilde{f}$.

The Lie point symmetries of partial differential equations can be restated more simply if we introduce the concept of the prolonged group action, which acts on $X \times U^{(n)}$. Denote the action of a group element $g \in G$ at a point $(x, u) \in X \times U$ as $(\tilde{x}, \tilde{u}) = g \cdot (x, u)$. Then, we define the $n$-th order prolongation of $g$ at the point $(x, u^{(n)}) \in X \times U^{(n)}$ as $\mathrm{pr}^{(n)} g \cdot (x, u^{(n)}) = (\tilde{x}, \tilde{u}^{(n)})$, where $\tilde{u}^{(n)}$ consists of all derivatives of $\tilde{u}$ with respect to $\tilde{x}$ up to order $n$. $G$ is a symmetry group of $F(x, u^{(n)}) = 0$ means that for any solution $u=f(x)$ and any group element $g \in G$, $F(\mathrm{pr}^{(n)} g \cdot (x, u^{(n)})) = 0$ holds.

\paragraph{Infinitesimal criteria.}

Suppose the Lie group $G$ corresponds to the Lie algebra $\mathfrak{g}$, which can be associated via the exponential map $\exp: \mathfrak{g} \rightarrow G$. The infinitesimal group action $\mathbf{v} \in \mathfrak{g}$ at the point $(x, u) \in X \times U$ is defined as $\left. \mathbf{v} \right|_{(x, u)} = \left. \frac{\mathrm{d}}{\mathrm{d} \epsilon} \right|_{\epsilon=0} \left[ \exp(\epsilon \mathbf{v}) \cdot (x, u) \right]$. Note that $\mathbf{v}$ is expressed in terms of the partial differential operator $\nabla$ as its special basis, which indicates that it can directly act on functions defined on $X \times U$. Taking the $\mathrm{SO}(2)$ group $\epsilon \cdot (x, u) = (x \cos \epsilon - u \sin \epsilon, x \sin \epsilon + u \cos \epsilon)$ as an example, its infinitesimal group action is $\left. \mathbf{v} \right|_{(x, u)} = -u \frac{\partial}{\partial x} + x \frac{\partial}{\partial u}$.

Similarly, we define the $n$-th order prolongation of $\mathbf{v} \in \mathfrak{g}$ at the point $(x, u^{(n)}) \in X \times U^{(n)}$ as $\left. \mathrm{pr}^{(n)} \mathbf{v} \right|_{(x, u^{(n)})} = \left. \frac{\mathrm{d}}{\mathrm{d} \epsilon} \right|_{\epsilon=0} \left\{ \mathrm{pr}^{(n)} \left[ \exp(\epsilon \mathbf{v}) \right] \cdot (x, u^{(n)}) \right\}$. Then, according to Theorem 2.31 in the textbook \citep{olver1993applications}, $G$ is a symmetry group of $F(x, u^{(n)}) = 0$ if, for every $\mathbf{v} \in \mathfrak{g}$, $\mathrm{pr}^{(n)} \mathbf{v} \left[ F(x, u^{(n)}) \right] = 0$ whenever $F(x, u^{(n)}) = 0$.

\section{Method}

\begin{figure}[h]
	\centering
	\includegraphics[width=\textwidth]{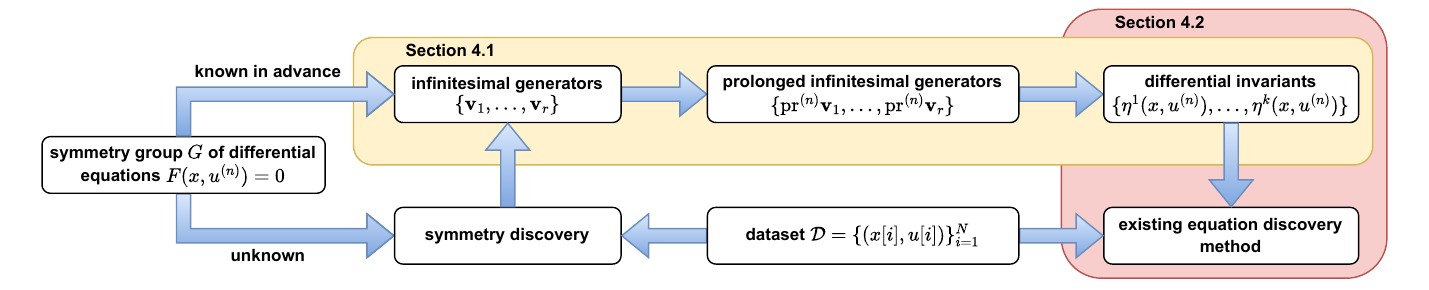}
	
	\caption{Pipeline of our differential invariant-based equation discovery method.}
	\label{fig:method}
\end{figure}

In short, we explore the use of prior knowledge about the symmetry group $G$ to guide the discovery of governing PDEs $F(x, u^{(n)}) = 0$ from the dataset $\mathcal{D} = \{(x[i], u[i])\}_{i=1}^N$. In Section \ref{sec:method1}, we prolong the infinitesimal generators of the symmetry group and compute the corresponding differential invariants. In Section \ref{sec:method2}, we discuss integrating differential invariants with existing equation discovery methods and provide a proposition to demonstrate that our approach is both correct and complete. In Section \ref{sec:method3}, we take SINDy \citep{brunton2016discovering} as an example to showcase the theoretical advantages of our method over other symmetry-guided equation discovery approaches, such as EquivSINDy-c and EquivSINDy-r \citep{yang2024symmetry}. Figure \ref{fig:method} provides an intuitive summary of our differential invariant-based equation discovery pipeline.

\subsection{Calculation of differential invariants}
\label{sec:method1}

Differential invariants refer to quantities that remain unchanged under the action of a prolonged group. Definition 2.51 in the textbook \citep{olver1993applications} provides a formal definition of differential invariants, which we briefly restate as follows.
\begin{definition}
	Let $G$ be a Lie group acting on $X \times U$. An $n$-th order differential invariant of $G$ is a smooth function $\eta: X \times U^{(n)} \rightarrow \mathbb{R}$ such that $\eta$ is an invariant under the prolonged group action $\mathrm{pr}^{(n)} G$:
	\begin{equation}
		\label{eq:di}
		\forall g \in G, (x, u^{(n)}) \in X \times U^{(n)}: \quad \eta(\mathrm{pr}^{(n)} g \cdot (x, u^{(n)})) = \eta(x, u^{(n)}).
	\end{equation}
\end{definition}
We now discuss how to find the differential invariants of a Lie group $G$. This problem can be formalized as follows: given the infinitesimal generators $\{\mathbf{v}_1, \dots, \mathbf{v}_r\}$ of the Lie group $G$, we seek a complete set of functionally independent $n$-th order differential invariants $\{ \eta^1(x, u^{(n)}), \dots, \eta^k(x, u^{(n)}) \}$ for $\mathrm{pr}^{(n)} G$ (functionally independent: they cannot be expressed as combinations of each other).

The first thing we need to do is derive the $n$-th order prolongation $\{ \mathrm{pr}^{(n)} \mathbf{v}_1, \dots, \mathrm{pr}^{(n)} \mathbf{v}_r \}$ of the infinitesimal generators. Consider an infinitesimal group action on $X \times U = \mathbb{R}^p \times \mathbb{R}^q$ in the form:
\begin{equation}
	\label{eq:prolong1}
	\mathbf{v} = \sum_{i=1}^p \xi^i(x, u) \frac{\partial}{\partial x^i} + \sum_{\alpha=1}^q \phi_\alpha(x, u) \frac{\partial}{\partial u^\alpha}.
\end{equation}
Then, according to Theorem 2.36 in the textbook \citep{olver1993applications}, its $n$-th order prolongation is:
\begin{equation}
	\label{eq:prolong2}
	\mathrm{pr}^{(n)} \mathbf{v} = \mathbf{v} + \sum_{\alpha=1}^q \sum_J \phi_\alpha^J (x, u^{(n)}) \frac{\partial}{\partial u_J^\alpha},
\end{equation}
where the coefficients are determined by:
\begin{equation}
	\label{eq:prolong3}
	\phi_\alpha^J(x, u^{(n)}) = \mathrm{D}_J \left( \phi_\alpha - \sum_{i=1}^p \xi^i u_i^\alpha \right) + \sum_{i=1}^p \xi^i u_{J, i}^\alpha.
\end{equation}
Here, $J=(j_1, \dots, j_k)$ with $j_i = 1, \dots, p$ and $k = 1, \dots, n$, $u_i^\alpha = \frac{\partial u^\alpha}{\partial x^i}$, and $u_{J, i}^\alpha = \frac{\partial u_J^\alpha}{\partial x^i} = \frac{\partial^{k+1} u^\alpha}{\partial x^i \partial x^{j_1} \dots \partial x^{j_k}}$. Note that $\mathrm{D}_J$ denotes the total derivative. For a smooth function $P(x, u^{(n)})$, its relationship with partial derivatives is given by $\mathrm{D}_i P = \frac{\partial P}{\partial x^i} + \sum_{\alpha=1}^q \sum_J u_{J, i}^\alpha \frac{\partial P}{\partial u_J^\alpha}$. Taking the infinitesimal group action $\mathbf{v} = -u \frac{\partial}{\partial x} + x \frac{\partial}{\partial u}$ of the $\mathrm{SO}(2)$ group as an example, its first-order prolongation is $\mathrm{pr}^{(1)} \mathbf{v} = \mathbf{v} + \phi^x(x, u, u_x) \frac{\partial}{\partial u_x}$, where $\phi^x(x, u, u_x) = \mathrm{D}_x (x + u u_x) - u u_{xx} = 1 + u_x^2$.

Next, we derive the $n$-th order differential invariants based on the prolonged infinitesimal generators. According to the infinitesimal criteria introduced in Section \ref{sec:preliminary}, Equation (\ref{eq:di}) is equivalent to:
\begin{equation}
	\label{eq:di1}
	\mathrm{pr}^{(n)} \mathbf{v} \left[ \eta(x, u^{(n)}) \right] = \sum_{i=1}^p \xi^i(x, u) \frac{\partial \eta}{\partial x^i} + \sum_{\alpha=1}^q \phi_\alpha(x, u) \frac{\partial \eta}{\partial u^\alpha} + \sum_{\alpha=1}^q \sum_J \phi_\alpha^J (x, u^{(n)}) \frac{\partial \eta}{\partial u_J^\alpha}.
\end{equation}
Then, we construct the characteristic equations:
\begin{equation}
	\label{eq:di2}
	\frac{\mathrm{d} x^i}{\xi^i(x, u)} = \frac{\mathrm{d} u^\alpha}{\phi_\alpha(x,u)} = \frac{\mathrm{d} u_J^\alpha}{\phi_\alpha^J(x, u^{(n)})},
\end{equation}
for all $i = 1, \dots, p$, $\alpha = 1, \dots, q$, and $J = (j_1, \dots, j_k)$ with $j_i = 1, \dots, p$ and $k = 1, \dots, n$. The integration constants of the general solution to the characteristic equations yield the differential invariants:
\begin{equation}
	\label{eq:di3}
	\eta^1(x, u^{(n)}) = c_1, \dots, \eta^k(x, u^{(n)}) = c_k.
\end{equation}
In the case of multiple prolonged infinitesimal generators, we solve the corresponding characteristic equations jointly. Taking the $\mathrm{SO}(2)$ group as an example again, the first-order prolongation of its infinitesimal generator is $\mathrm{pr}^{(1)} \mathbf{v} = -u \frac{\partial}{\partial x} + x \frac{\partial}{\partial u} + (1 + u_x^2) \frac{\partial}{\partial u_x}$. We construct the characteristic equation as $\frac{\mathrm{d} x}{-u} = \frac{\mathrm{d} u}{x} = \frac{\mathrm{d} u_x}{1 + u_x^2}$. The constants obtained by integration are $\eta^1(x, u, u_x) = \sqrt{x^2 + u^2}$ and $\eta^2(x, u, u_x) = \frac{x u_x - u}{u u_x + x}$, which constitute the first-order differential invariants of the $\mathrm{SO}(2)$ group.

\subsection{Governing equation discovery based on differential invariants}
\label{sec:method2}

Existing equation discovery methods typically follow the paradigm of first specifying the equation skeleton and then optimizing the parameters. When manually specifying the equation skeleton, the challenge lies in selecting the relevant terms. Including too many irrelevant terms leads to excessive computational costs and reduced accuracy, while omitting key terms makes it theoretically impossible for the algorithm to achieve the correct solution. This limitation becomes even more pronounced in partial differential equation discovery, as compared to $X \times U$, $X \times U^{(n)}$ usually constitutes a much larger search space with more candidate terms to choose from.

Our method aims to use symmetry to guide the selection of relevant terms. We hope that this selection approach, while respecting symmetry, can provide a relatively concise search space without losing expressive power. Proposition 2.56 in the textbook \citep{olver1993applications} provides the inspiration, which we briefly restate as follows.

\begin{proposition}
	\label{pro:di}
	Let $G$ be a Lie group acting on $X \times U$, and $\eta^1(x, u^{(n)}), \dots, \eta^k(x, u^{(n)})$ be a complete set of functionally independent $n$-th order differential invariants. An $n$-th order differential equation $F(x, u^{(n)}) = 0$ admits $G$ as a symmetry group if and only if there is an equivalent equation
	\begin{equation}
		\widetilde{F}(\eta^1(x, u^{(n)}), \dots, \eta^k(x, u^{(n)})) = 0
	\end{equation}
	involving only the differential invariants of $G$.
\end{proposition}

Therefore, we first use the procedure in Section \ref{sec:method1} to compute differential invariants based on the symmetry group, which serve as all the relevant terms. Then, we can choose any existing equation discovery method \citep{brunton2016discovering,champion2019data,messenger2021weak,biggio2021neural} to explicitly solve for $\widetilde{F}$. Our approach does not interfere with the core of these methods, except for providing the selection of relevant terms, which means it is plug-and-play. Proposition \ref{pro:di} theoretically guarantees that this substitution approach strictly adheres to the symmetry prior while ensuring that the equation skeleton is not missing potential solutions due to the omission of relevant terms. When the symmetry is unknown, we can first employ symmetry discovery methods \citep{yang2023latent,ko2024learning,shaw2024symmetry} to obtain infinitesimal generators from the data and then implement the aforementioned equation discovery process.

Note that we do not need to exhaustively provide all infinitesimal generators of the symmetry group. In most cases, we might miss some infinitesimal generators due to reasons such as errors in symmetry detection, but this does not affect the correctness of the equation discovery results. This is because if a Lie group $G$ is the symmetry group of a differential equation, so is any subgroup $\widetilde{G} \subseteq G$. In fact, each additional correct infinitesimal generator we provide reduces the complete set of functionally independent differential invariants, which leads to a smaller and more accurate search space for the governing equation. In Table~\ref{tab:di}, we use the Lie point symmetries of the KdV, KS, and Burgers equations mentioned by \citet{ko2024learning} as examples to demonstrate the complete set of functionally independent differential invariants corresponding to different numbers of infinitesimal generators.

\begin{table}[h]
	\caption{The complete set of functionally independent differential invariants corresponding to different numbers of provided infinitesimal generators. For detailed calculation steps, refer to Appendix \ref{sec:calculation1}.}
	\label{tab:di}
	\centering
	\resizebox{\textwidth}{!}{
	\begin{tabular}{ll}
		\toprule
		Provided infinitesimal generators & Complete set of functionally independent differential invariants \\
		\midrule
		$\emptyset$ & $\{t, x, u, u_t, u_x, u_{xx}, u_{xxx}, u_{xxxx}\}$ \\
		$\{\partial_x\}$ & $\{t, u, u_t, u_x, u_{xx}, u_{xxx}, u_{xxxx}\}$ \\
		$\{\partial_x, \partial_t\}$ & $\{u, u_t, u_x, u_{xx}, u_{xxx}, u_{xxxx}\}$ \\
		$\{\partial_x, \partial_t, t \partial_x + \partial_u\}$ & $\{u_t + u u_x, u_x, u_{xx}, u_{xxx}, u_{xxxx}\}$ \\
		\bottomrule
	\end{tabular}
	}
\end{table}

\subsection{Example algorithm: DI-SINDy}
\label{sec:method3}

\begin{algorithm}[h]
	\caption{DI-SINDy (SINDy based on Differential Invariants)}
	\label{alg:overall}
	\begin{algorithmic}
		\STATE {\bfseries Input:} Dataset $\mathcal{D} = \{(x[i], u[i])\}_{i=1}^N$, prolongation order $n$, infinitesimal generators of the symmetry group $V(\mathfrak{g}) = \{\mathbf{v}_1, \dots, \mathbf{v}_r\}$.
		\STATE {\bfseries Output:} Explicit governing equation $F(x, u^{(n)}) = 0$.
		\STATE {\bfseries Execute:}
		\STATE Estimate the derivatives of $u$ with respect to $x$ using the central difference method, resulting in the prolonged dataset $\mathrm{pr}^{(n)} \mathcal{D} = \{(x[i], u^{(n)}[i])\}_{i=1}^N$.
		\IF{$V(\mathfrak{g}) = \emptyset$}
		\STATE Use the method of symmetry discovery to obtain the infinitesimal generators $V(\mathfrak{g}) = \{\mathbf{v}_1, \dots, \mathbf{v}_r\}$ of the symmetry group from $\mathrm{pr}^{(n)} \mathcal{D}$.
		\ENDIF
		\STATE Derive the prolonged infinitesimal generators $\{ \mathrm{pr}^{(n)} \mathbf{v}_1, \dots, \mathrm{pr}^{(n)} \mathbf{v}_r \}$ according to Equations (\ref{eq:prolong1}) to (\ref{eq:prolong3}).
		\STATE Compute differential invariants $\{\eta^1(x, u^{(n)}), \dots, \eta^k(x, u^{(n)}) \}$ according to Equations (\ref{eq:di1}) to (\ref{eq:di3}).
		For the equation skeleton $\eta^k(x, u^{(n)}) = W \Theta(\eta^1(x, u^{(n)}), \dots, \eta^{k-1}(x, u^{(n)}))$, optimize the coefficient matrix $W$ using SINDy based on $\mathrm{pr}^{(n)} \mathcal{D}$.
		\STATE {\bfseries Return} $F(x, u^{(n)}) = \eta^k(x, u^{(n)}) - W \Theta(\eta^1(x, u^{(n)}), \dots, \eta^{k-1}(x, u^{(n)})) = 0$.
	\end{algorithmic}
\end{algorithm}

Now our method can be summarized as follows. First, we use symmetry discovery methods to obtain infinitesimal generators from the dataset if the symmetries are not known a priori. Then, we derive the prolonged infinitesimal generators and compute the differential invariants based on them. Finally, we select the relevant terms of the equation skeleton from the differential invariants and employ existing equation discovery methods to obtain the explicit governing equation. Taking SINDy \citep{brunton2016discovering} based on Differential Invariants (DI-SINDy) as an example, we outline the overall workflow in Algorithm \ref{alg:overall}.

The EquivSINDy-c and EquivSINDy-r methods proposed by \citet{yang2024symmetry} also attempt to use symmetry to guide SINDy in discovering governing equations of the form $h(x) = W \Theta(x)$. However, for EquivSINDy-c, it cannot handle nonlinear cases, and Proposition 4.2 in the original paper \citep{yang2024symmetry} specifies that $\Theta(x)$ can only be chosen as polynomials. Additionally, the constrained parameter space of $W$ reduces the expressive power of the equation skeleton. On the other hand, the necessity and sufficiency of Proposition \ref{pro:di} in this paper guarantee that DI-SINDy’s skeleton can fully express all equations satisfying the symmetry, and $\Theta(x)$ can be freely selected, thereby addressing the limitations of EquivSINDy-c. Compared to EquivSINDy-r, which incorporates symmetry loss as a regularization term into SINDy's loss function, DI-SINDy ensures that the equation skeleton strictly adheres to symmetry without requiring hyperparameter tuning for regularization coefficients. Overall, DI-SINDy holds significant theoretical advantages over related works, thanks to its intrinsic ability to ``losslessly'' compress the equation search space based on symmetry.

\section{Experiment}

\subsection{Experimental setup}
\label{sec:setup}

We evaluate our method using the Korteweg-de Vries (KdV) equation, the Kuramoto-Shivashinsky (KS) equation, the Burgers equation, and the nKdV equation from \citet{ko2024learning}. In Table \ref{tab:pde}, we present their explicit equations, the infinitesimal generators of their symmetry groups, and the corresponding differential invariants (detailed calculation steps are provided in Appendix \ref{sec:calculation}), where the prolongation order is specified as fourth-order. We assume the symmetries are known a priori, and the experimental task is to automatically discover the governing equations from the generated data. The infinitesimal generators provided here are all sufficiently simple to be easily obtained by existing symmetry discovery methods. We provide the data generation process in Appendix \ref{sec:data}.

\begin{table}[h]
	\caption{Explicit expressions, infinitesimal generators of symmetry groups, and corresponding differential invariants for the KdV, KS, Burgers, and nKdV equations \citep{ko2024learning}.}
	\label{tab:pde}
	\centering
	\resizebox{\textwidth}{!}{
	\begin{tabular}{llll}
		\toprule
		Name & Equation & Infinitesimal generators & Differential invariants \\
		\midrule
		KdV & $u_t + u u_x + u_{xxx} = 0$ & \multirow{3}{*}{$\{\frac{\partial}{\partial x}, \frac{\partial}{\partial t}, t \frac{\partial}{\partial x} + \frac{\partial}{\partial u}\}$} & \multirow{3}{*}{$\{u_t + u u_x, u_x, u_{xx}, u_{xxx}, u_{xxxx}\}$} \\
		KS & $u_t + u_{xx} + u_{xxxx} + u u_x = 0$ &  &  \\
		Burgers & $u_t + u u_x - \nu u_{xx} = 0$ &  &  \\
		\midrule
		nKdV & $e^{-\frac{t}{t_0}} u_t + u u_x + u_{xxx} = 0$ & $\{\frac{\partial}{\partial x}, e^{-\frac{t}{t_0}} \frac{\partial}{\partial t}, t_0 (e^\frac{t}{t_0} - 1) \frac{\partial}{\partial x} + \frac{\partial}{\partial u}\}$ & $\{e^{-\frac{t}{t_0}} u_t + u u_x, u_x, u_{xx}, u_{xxx}, u_{xxxx}\}$ \\
		\bottomrule
	\end{tabular}
	}
\end{table}

Taking DI-SINDy presented in Algorithm \ref{alg:overall} as an example, we compare it with SINDy \citep{brunton2016discovering} and EquivSINDy-r \citep{yang2024symmetry}. The Lie point symmetry of PDEs is typically nonlinear, which renders EquivSINDy-c inapplicable—hence we exclude it from the comparison. The idea behind EquivSINDy-r is to incorporate the infinitesimal criterion of the symmetry group as a regularization term into the objective function of SINDy, thereby softly constraining the equation skeleton to adhere to the symmetry. The original paper \citep{yang2024symmetry} only provides the form of the regularization term for ODE cases. To extend it to PDE scenarios for comparison, we adopt the infinitesimal criterion of Lie point symmetry introduced in Section \ref{sec:preliminary} as the regularization term:
\begin{equation}
	\mathcal{L}_{symm} = \mathbb{E}_{x, u} \left\{ \sum_{\mathbf{v} \in V(\mathfrak{g})} \left\| \mathrm{pr}^{(n)} \mathbf{v} \left[ F(x, u^{(n)}) \right] \right\|^2 \right\},
\end{equation}
where $V(\mathfrak{g})$ is the set of infinitesimal generators of the symmetry group, and $F$ represents the equation skeleton of SINDy. Then, the overall objective function of EquivSINDy-r is:
\begin{equation}
	\mathcal{L}_{total} = \mathcal{L}_{SINDy} + \lambda \cdot \mathcal{L}_{symm}.
\end{equation}
For a comprehensive comparison, we will traverse the regularization weight hyperparameter $\lambda=\{10^{-3}, 10^{-2}, 10^{-1}\}$.

As described in Algorithm \ref{alg:overall}, the relevant terms of DI-SINDy are selected as the set of differential invariants shown in Table \ref{tab:pde}, and the function library $\Theta$ is specified as linear terms. For SINDy and EquivSINDy-r, we define the equation skeleton of the KdV, KS, and Burgers equations as $u_t = W \Theta(u, u_x, u_{xx}, u_{xxx}, u_{xxxx})$, and the equation skeleton of the nKdV equation as $e^{-\frac{t}{t_0}} u_t = W \Theta(u, u_x, u_{xx}, u_{xxx}, u_{xxxx})$, where $\Theta$ contains terms up to second order. It can be observed that the baseline methods require strong prior assumptions about the equation skeleton during the experimental preparation phase, even though we have manually specified relatively simple forms for them that include the ground truth. More implementation details can be found in Appendix \ref{sec:implementation}.

\subsection{Quantitative metrics and result analysis}

After training with SINDy and its variant methods, we get explicit equations such as $u_t = W \Theta(u, u_x, \dots)$ (for KdV, KS, and Burgers equations) or $e^{-\frac{t}{t_0}} u_t = W \Theta(u, u_x, \dots)$ (for the nKdV equation). In practice, the coefficient matrix is obtained via element-wise multiplication $W = C \odot M$, where $C$ represents the values of each term's coefficient, and the binary mask matrix $M$ indicates whether each term is retained ($1$ for retained, $0$ for discarded). We follow the quantitative metrics introduced by \citet{yang2024symmetry}, which we restate as follows. We consider the discovery of an equation successful if the retained terms in the final result are correct and complete (formally, $M=M^*$, where $M^*$ is the ground truth of the binary mask matrix). We run each experiment $50$ times and calculate its \textbf{success rate}, which is the most important quantitative metric for explicit equation discovery, as it reflects whether the model can correctly identify the interaction relationships between variables. Furthermore, we use the RMSE of the coefficient matrix, $\sqrt{\frac{1}{n} \sum_{i=1}^n \|W - W^*\|^2}$, to evaluate the accuracy of equation discovery, where $n$ is the number of runs, and $W^*$ is the ground truth of the coefficient matrix. We report \textbf{RMSE (successful)} and \textbf{RMSE (all)}, which represent the RMSE for successful runs and all runs, respectively.

\begin{table}[h]
	\caption{Success rates and RMSE of different equation discovery methods for the KdV, KS, Burgers, and nKdV equations. All experimental results are averaged over $50$ runs. RMSE is presented in the format of mean $\pm$ std.}
	\label{tab:result}
	\centering
	\resizebox{\textwidth}{!}{
	\begin{tabular}{lllll}
		\toprule
		Name & Method & Success rate ($\uparrow$) & RMSE (successful) ($\downarrow$) & RMSE (all) ($\downarrow$) \\
		\midrule
		\multirow{5}{*}{KdV} & SINDy & $72\%$ & $(2.24 \pm 0.51) \times 10^{-1}$ & $(4.42 \pm 3.51) \times 10^{-1}$ \\
		& EquivSINDy-r ($\lambda=10^{-3}$) & $72\%$ & $(2.23 \pm 0.51) \times 10^{-1}$ & $(4.41 \pm 3.51) \times 10^{-1}$ \\
		& EquivSINDy-r ($\lambda=10^{-2}$) & $74\%$ & $(2.18 \pm 0.50) \times 10^{-1}$ & $(9.28 \pm 14.01) \times 10^{-2}$ \\
		& EquivSINDy-r ($\lambda=10^{-1}$) & $82\%$ & $(1.66 \pm 0.37) \times 10^{-1}$ & $(3.16 \pm 3.22) \times 10^{-1}$ \\
		& DI-SINDy (Ours) & $\mathbf{100\%}$ & $\mathbf{(2.71 \pm 2.44) \times 10^{-2}}$ & $\mathbf{(2.71 \pm 2.44) \times 10^{-2}}$ \\
		\midrule
		\multirow{5}{*}{KS} & SINDy & $0\%$ & N/A & $1.00 \pm 0.00$ \\
		& EquivSINDy-r ($\lambda=10^{-3}$) & $0\%$ & N/A & $1.00 \pm 0.00$ \\
		& EquivSINDy-r ($\lambda=10^{-2}$) & $0\%$ & N/A & $1.00 \pm 0.00$ \\
		& EquivSINDy-r ($\lambda=10^{-1}$) & $0\%$ & N/A & $1.00 \pm 0.00$ \\
		& DI-SINDy (Ours) & $\mathbf{100\%}$ & $\mathbf{(6.18 \pm 0.37) \times 10^{-2}}$ & $\mathbf{(6.18 \pm 0.37) \times 10^{-2}}$ \\
		\midrule
		\multirow{5}{*}{Burgers} & SINDy & $4\%$ & $(2.11 \pm 0.14) \times 10^{-2}$ & $(1.52 \pm 2.34) \times 10^{-1}$ \\
		& EquivSINDy-r ($\lambda=10^{-3}$) & $16\%$ & $(2.59 \pm 0.42) \times 10^{-2}$ & $(1.86 \pm 4.12) \times 10^{-1}$ \\
		& EquivSINDy-r ($\lambda=10^{-2}$) & $68\%$ & $(8.06 \pm 3.38) \times 10^{-3}$ & $(9.78 \pm 38.08) \times 10^{-2}$ \\
		& EquivSINDy-r ($\lambda=10^{-1}$) & $78\%$ & $(9.68 \pm 3.89) \times 10^{-4}$ & $(7.03 \pm 35.62) \times 10^{-2}$ \\
		& DI-SINDy (Ours) & $\mathbf{98\%}$ & $\mathbf{(2.66 \pm 1.32) \times 10^{-4}}$ & $\mathbf{(4.02 \pm 9.62) \times 10^{-4}}$ \\
		\midrule
		\multirow{5}{*}{nKdV} & SINDy & $20\%$ & $(3.77 \pm 0.14) \times 10^{-1}$ & $(8.75 \pm 2.49) \times 10^{-1}$ \\
		& EquivSINDy-r ($\lambda=10^{-3}$) & $20\%$ & $(3.76 \pm 0.14) \times 10^{-1}$ & $(8.75 \pm 2.50) \times 10^{-1}$ \\
		& EquivSINDy-r ($\lambda=10^{-2}$) & $22\%$ & $(3.62 \pm 0.13) \times 10^{-1}$ & $(8.60 \pm 2.64) \times 10^{-1}$ \\
		& EquivSINDy-r ($\lambda=10^{-1}$) & $44\%$ & $(2.70 \pm 0.19) \times 10^{-1}$ & $(6.79 \pm 3.63) \times 10^{-1}$ \\
		& DI-SINDy (Ours) & $\mathbf{100\%}$ & $\mathbf{(5.05 \pm 3.84) \times 10^{-2}}$ & $\mathbf{(5.05 \pm 3.84) \times 10^{-2}}$ \\
		\bottomrule
	\end{tabular}
	}
\end{table}

The success rates and RMSE of different equation discovery methods are presented in Table \ref{tab:result}. For the KdV, Burgers, and nKdV equations, EquivSINDy-r, with its soft symmetry constraints, significantly improves both the success rate and accuracy compared to SINDy, while our DI-SINDy further increases the success rate to nearly $100\%$. Notably, both SINDy and EquivSINDy-r fail for the KS equation, as the KS equation involves a fourth-order derivative term, making finite difference methods prone to large errors in the presence of noise. In contrast, DI-SINDy, benefiting from a smaller search space, can still accurately identify the correct equation form, demonstrating stronger robustness.

Beyond quantitative advantages, as discussed in Section \ref{sec:setup}, DI-SINDy employs differential invariants as candidate terms, unlike SINDy and EquivSINDy-r, which rely on manually specified equation skeletons (e.g., for the nKdV equation, the term $e^{-\frac{t}{t_0}} u_t$ is difficult to guess, whereas differential invariants naturally guide its inclusion). Additionally, the performance of EquivSINDy-r is sensitive to the regularization weight $\lambda$, while DI-SINDy eliminates the need for hyperparameter tuning.

\begin{figure}[htbp]
	\centering
	\begin{minipage}[b]{0.49\textwidth}
		\includegraphics[width=\textwidth]{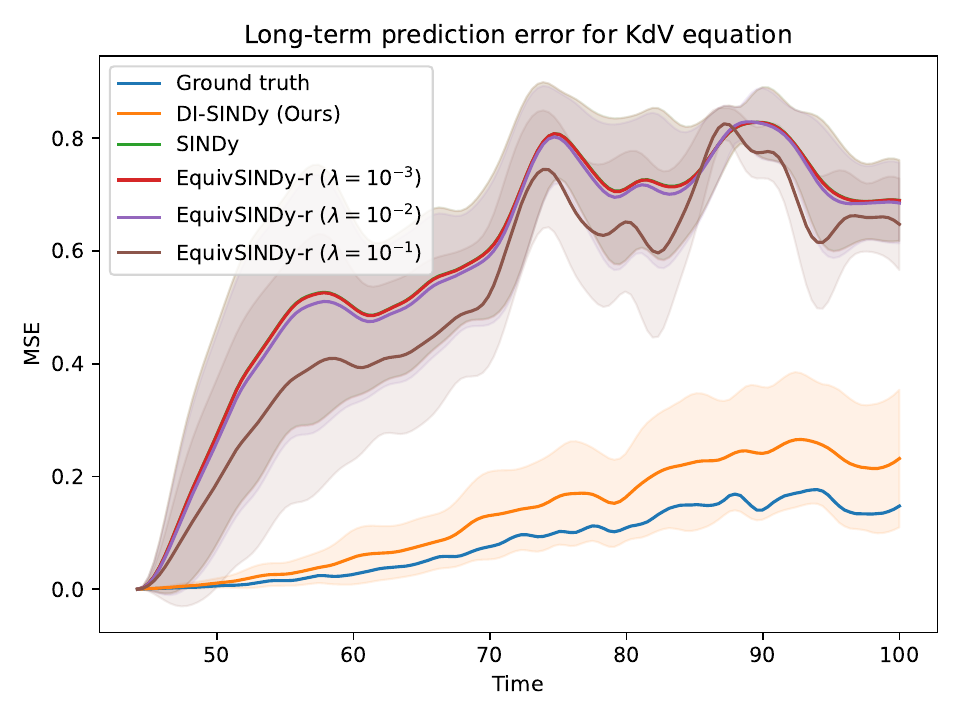}
	\end{minipage}
	\hfill
	\begin{minipage}[b]{0.49\textwidth}
		\includegraphics[width=\textwidth]{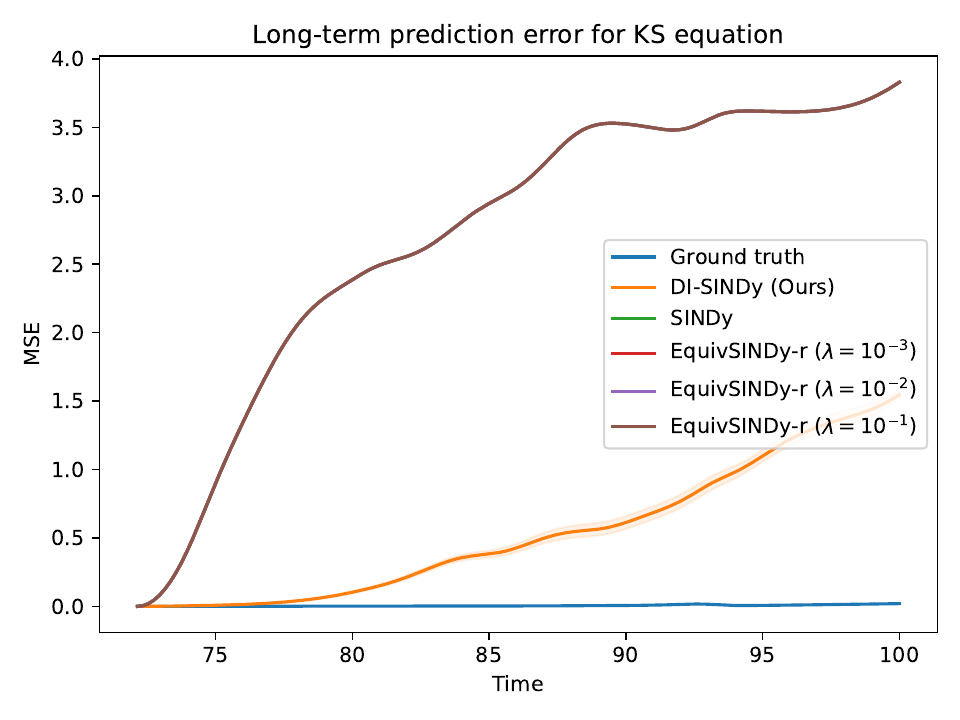}
	\end{minipage}
	\\
	\begin{minipage}[b]{0.49\textwidth}
		\includegraphics[width=\textwidth]{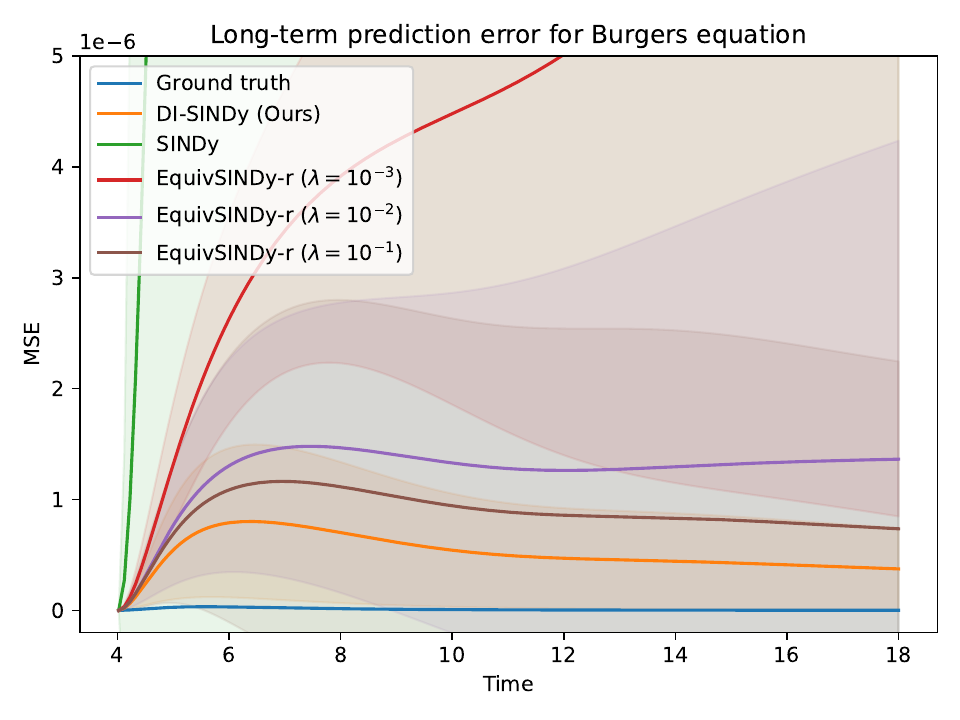}
	\end{minipage}
	\hfill
	\begin{minipage}[b]{0.49\textwidth}
		\includegraphics[width=\textwidth]{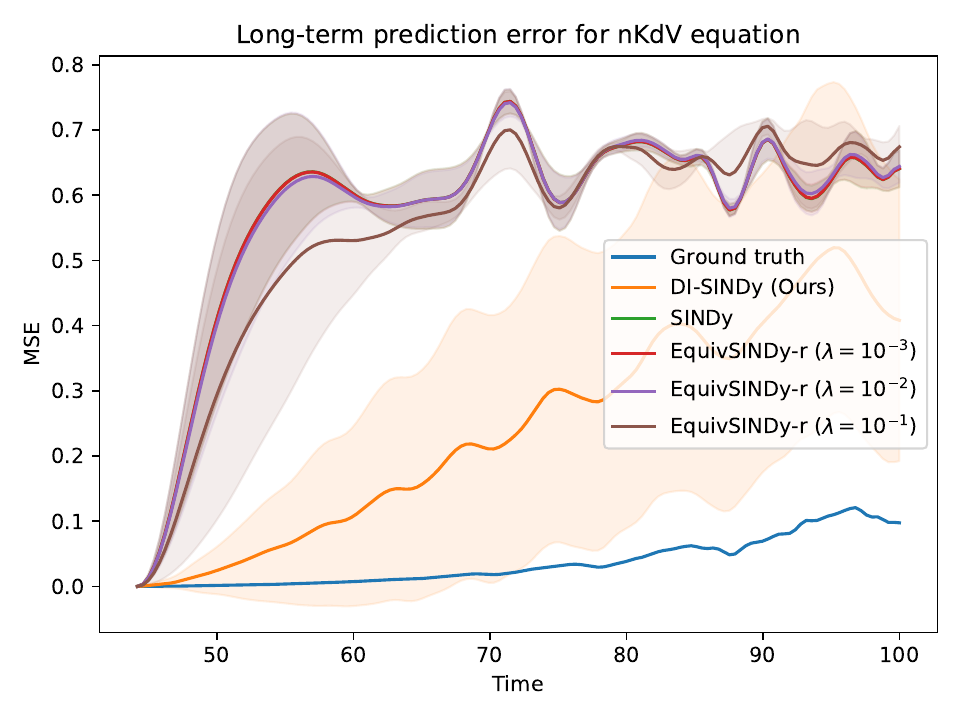}
	\end{minipage}
	
	\caption{Long-term prediction errors of different equation discovery methods for the KdV, KS, Burgers, and nKdV equations. The MSE at each time step is averaged over $4$ initial conditions and $50$ runs, with the shaded area representing the standard deviation.}
	\label{fig:ltp}
\end{figure}

We further numerically integrate the discovered explicit equations for the $4$ initial conditions in the test dataset and calculate the MSE against their corresponding true trajectories, which we refer to as the \textbf{long-term prediction error}. In Figure \ref{fig:ltp}, we visualize the long-term prediction errors of all methods for the KdV, KS, Burgers, and nKdV equations as a function of the integration time steps. We use the ground-truth equation form as the benchmark (blue lines), for which the long-term prediction error primarily stems from finite differences and numerical integration. For the KdV and nKdV equations, the error curves of SINDy and EquivSINDy-r ($\lambda=10^{-3}$) almost overlap, while for the KS equation, the error curves of SINDy and EquivSINDy-r with all $\lambda$ values nearly coincide. This is due to the minimal differences in their discovered explicit equations, which can be verified by the numerical results in Table \ref{tab:result}. For all PDEs, our DI-SINDy achieves significantly lower long-term prediction errors than baselines, further validating the accuracy of its equation discovery results.

\section{Conclusion and limitation}
\label{sec:conclusion}

Overall, our method addresses several pain points in existing equation discovery approaches. For the large search space of PDEs, most methods struggle to identify the correct relevant terms, whereas we overcome this limitation by employing differential invariants. The necessity and sufficiency of Proposition \ref{pro:di} show that our method neither loses expressiveness like symmetry-constrained approaches such as EquivSINDy-c, nor violates symmetry principles like regularization-based methods such as EquivSINDy-r. However, our approach relies on prior knowledge of the correct symmetry group. Although we claim that our approach can be combined with data-driven symmetry discovery techniques, inaccuracies in automatically identified symmetries may affect the precision of equation discovery results. As more robust symmetry discovery methods emerge in the future, we believe this limitation will be resolved.

\newpage
\bibliographystyle{plainnat}
\bibliography{bib/references}


\newpage
\appendix
\section{Applications of symmetry}
\label{sec:application}

Symmetry plays an important role in both traditional mathematical physics problems and the field of deep learning. For the mathematical solution of differential equations, symmetry can guide variable substitutions to reduce their order \citep{olver1993applications,mclachlan1995numerical,ibragimov1999elementary,hydon2000symmetry,bluman2008symmetry,bluman2010applications}. In recent years, equivariant networks have incorporated symmetry into network architectures, significantly improving performance and generalization in specific scientific and computer vision tasks \citep{zaheer2017deep,weiler2018learning,weiler20183d,kondor2018generalization,wang2020incorporating,finzi2021practical,satorras2021n,ruhe2023clifford}. Additionally, symmetry has been introduced into Physics-Informed Neural Networks (PINNs) or used to guide data augmentation to enhance the accuracy of neural PDE solvers \citep{arora2024invariant,lagrave2022equivariant,shumaylov2024lie,li2022physics,zhang2023enforcing,wang2025generalized,akhound2023lie,brandstetter2022lie}. Notably, our goal is to discover explicit equations rather than using PINNs to learn the evolution process of PDEs, which means the problem we focus on differs from that of neural PDE solvers.

\section{Example}
\label{sec:example}

We take the KdV equation $u_t + u u_x + u_{xxx} = 0$ as an example to intuitively understand the concepts introduced in Section \ref{sec:preliminary}. In this case, the independent variables are $(x, t) \in X = \mathbb{R}^2$, and the dependent variable is $u \in U = \mathbb{R}$. Consider the group $G$ acting on $X \times U$, which includes three types of group actions:
\begin{equation}
	\begin{cases}
		\epsilon_1 \cdot (x, t, u) = (x + \epsilon_1, t, u), \\
		\epsilon_2 \cdot (x, t, u) = (x, t + \epsilon_2, u), \\
		\epsilon_3 \cdot (x, t, u) = (x + \epsilon_3 t, t, u + \epsilon_3).
	\end{cases}
\end{equation}
According to the definition $\left. \mathbf{v} \right|_{(x, u)} = \left. \frac{\mathrm{d}}{\mathrm{d} \epsilon} \right|_{\epsilon=0} \left[ \exp(\epsilon \mathbf{v}) \cdot (x, u) \right]$, the infinitesimal generators are:
\begin{equation}
	\begin{cases}
		\mathbf{v}_1 = \frac{\partial}{\partial x}, \\
		\mathbf{v}_2 = \frac{\partial}{\partial t}, \\
		\mathbf{v}_3 = t \frac{\partial}{\partial x} + \frac{\partial}{\partial u}.
	\end{cases}
\end{equation}
Assuming $u = f(x, t)$ is a solution to the KdV equation, then under the aforementioned three types of group actions, the graph $\Gamma_f = \{(x, t, f(x, t)): (x, t) \in X\}$ is transformed into the graphs of the following three functions, respectively:
\begin{equation}
	\begin{cases}
		u^{(1)} = f(x - \epsilon_1, t), \\
		u^{(2)} = f(x, t - \epsilon_2), \\
		u^{(3)} = f(x - \epsilon_3 t, t) + \epsilon_3.
	\end{cases}
\end{equation}
It is easy to verify that if $u=f(x, t)$ satisfies the KdV equation, then $u^{(1)}, u^{(2)}, u^{(3)}$ are also solutions of the equation. Therefore, we call $G$ the symmetry group of the KdV equation.

Note that $u^{(3)}_t = -\epsilon_3 f_x(x - \epsilon_3 t, t) + f_t(x - \epsilon_3 t, t)$. The forms of the other transformed derivatives remain unchanged. Then, we can provide the prolongation of group actions:
\begin{equation}
	\begin{cases}
		\mathrm{pr}^{(n)} \epsilon_1 \cdot (x, t, u, u_t, u_x, \dots) = (x + \epsilon_1, t, u, u_t, u_x, \dots), \\
		\mathrm{pr}^{(n)} \epsilon_2 \cdot (x, t, u, u_t, u_x, \dots) = (x, t + \epsilon_2, u, u_t, u_x, \dots), \\
		\mathrm{pr}^{(n)} \epsilon_3 \cdot (x, t, u, u_t, u_x, \dots) = (x + \epsilon_3 t, t, u + \epsilon_3, -\epsilon_3 u_x + u_t, u_x, \dots).
	\end{cases}
\end{equation}
According to the definition $\left. \mathrm{pr}^{(n)} \mathbf{v} \right|_{(x, u^{(n)})} = \left. \frac{\mathrm{d}}{\mathrm{d} \epsilon} \right|_{\epsilon=0} \left\{ \mathrm{pr}^{(n)} \left[ \exp(\epsilon \mathbf{v}) \right] \cdot (x, u^{(n)}) \right\}$, the prolongation of the infinitesimal generators are:
\begin{equation}
	\begin{cases}
		\mathrm{pr}^{(n)} \mathbf{v}_1 = \frac{\partial}{\partial x}, \\
		\mathrm{pr}^{(n)} \mathbf{v}_2 = \frac{\partial}{\partial t}, \\
		\mathrm{pr}^{(n)} \mathbf{v}_3 = t \frac{\partial}{\partial x} + \frac{\partial}{\partial u} - u_x \frac{\partial}{\partial u_t}.
	\end{cases}
\end{equation}
Then, we can observe that the infinitesimal criteria $\mathrm{pr}^{(n)} \mathbf{v}_i (u_t + u u_x + u_{xxx}) = 0$ hold for $i=1,2,3$.

\section{Detailed calculation steps of differential invariants}
\label{sec:calculation}

Consider the case where $X \times U = \mathbb{R}^2 \times \mathbb{R}$, with $(x, t) \in X$ as the independent variables and $u \in U$ as the dependent variable. We specify the highest prolongation order as $n=4$, so the initial search space consists of the terms $\{t, x, u, u_t, u_x, u_{xx}, u_{xxx}, u_{xxxx}\}$ (for simplicity, we assume the dynamical system is first-order, meaning the highest-order partial derivative of $u$ with respect to $t$ is first-order).

\subsection{KdV, KS, and Burgers equations}
\label{sec:calculation1}

As shown in Table \ref{tab:pde}, the infinitesimal generators of the symmetry groups for the KdV, KS, and Burgers equations are:
\begin{equation}
	\mathbf{v}_1 = \frac{\partial}{\partial x}, \quad \mathbf{v}_2 = \frac{\partial}{\partial t}, \quad \mathbf{v}_3 = t \frac{\partial}{\partial x} + \frac{\partial}{\partial u}.
\end{equation}
We first compute their fourth-order prolongations. For $\mathrm{pr}^{(4)} \mathbf{v}_1$, we calculate its coefficients from Equation (\ref{eq:prolong3}):
\begin{equation}
	\begin{cases}
		\phi^t = \mathrm{D}_t (-u_x) + u_{tx} = 0, \\
		\phi^x = \mathrm{D}_x (-u_x) + u_{xx} = 0, \\ 
		\phi^{xx} = \mathrm{D}_{xx} (-u_x) + u_{xxx} = 0, \\
		\phi^{xxx} = \mathrm{D}_{xxx} (-u_x) + u_{xxxx} = 0, \\ 
		\phi^{xxxx} = \mathrm{D}_{xxxx} (-u_x) + u_{xxxxx} = 0.
	\end{cases}
\end{equation}
Therefore, we have:
\begin{equation}
	\label{eq:tmp}
	\mathrm{pr}^{(4)} \mathbf{v}_1 = \mathbf{v}_1 = \frac{\partial}{\partial x}.
\end{equation}
Similarly, it can be obtained that:
\begin{equation}
	\mathrm{pr}^{(4)} \mathbf{v}_2 = \mathbf{v}_2 = \frac{\partial}{\partial t}.
\end{equation}
The coefficients of $\mathrm{pr}^{(4)} \mathbf{v}_3$ are calculated as follows:
\begin{equation}
	\begin{cases}
		\phi^t = \mathrm{D}_t (1-t u_x) + t u_{tx} = -u_x, \\
		\phi^x = \mathrm{D}_x (1-t u_x) + t u_{xx} = 0, \\ 
		\phi^{xx} = \mathrm{D}_{xx} (1-t u_x) + t u_{xxx} = 0, \\
		\phi^{xxx} = \mathrm{D}_{xxx} (1-t u_x) + t u_{xxxx} = 0, \\ 
		\phi^{xxxx} = \mathrm{D}_{xxxx} (1-t u_x) + t u_{xxxxx} = 0.
	\end{cases}
\end{equation}
This means:
\begin{equation}
	\mathrm{pr}^{(4)} \mathbf{v}_3 = \mathbf{v}_3 - u_x \frac{\partial}{\partial u_t} = t \frac{\partial}{\partial x} + \frac{\partial}{\partial u} - u_x \frac{\partial}{\partial u_t} .
\end{equation}
Substitute $\mathrm{pr}^{(4)} \mathbf{v}_1$ and $\mathrm{pr}^{(4)} \mathbf{v}_2$ into Equation (\ref{eq:di1}):
\begin{equation}
	\label{eq:tmp2}
	\frac{\partial \eta}{\partial x} = \frac{\partial \eta}{\partial t} = 0.
\end{equation}
Therefore, the differential invariants do not contain the terms $x$ and $t$. The search space can be narrowed down to $\{u, u_t, u_x, u_{xx}, u_{xxx}, u_{xxxx}\}$.

For $\mathrm{pr}^{(4)} \mathbf{v}_3$, we can construct the characteristic equation as shown in Equation (\ref{eq:di2}) (Note that the term $x$ has already been excluded, so the $t \frac{\partial}{\partial x}$ in $\mathrm{pr}^{(4)} \mathbf{v}_3$ can be ignored):
\begin{equation}
	\mathrm{d} u = -\frac{\mathrm{d} u_t}{u_x}.
\end{equation}
By integrating it, we get:
\begin{equation}
	u_t + u u_x = c.
\end{equation}
Replacing $u$ and $u_t$ in the search space with the integration constant $u_t + u u_x$, we obtain the final differential invariants $\{u_t + u u_x, u_x, u_{xx}, u_{xxx}, u_{xxxx}\}$.

\subsection{nKdV equation}
\label{sec:calculation2}

The infinitesimal generators of the symmetry group for the nKdV equation are shown in Table \ref{tab:pde} as:
\begin{equation}
	\mathbf{v}_1 = \frac{\partial}{\partial x}, \quad \mathbf{v}_2 = e^{-\frac{t}{t_0}} \frac{\partial}{\partial t}, \quad \mathbf{v}_3 = t_0 (e^\frac{t}{t_0} - 1) \frac{\partial}{\partial x} + \frac{\partial}{\partial u}.
\end{equation}
The form of $\mathrm{pr}^{(4)} \mathbf{v}_1$ is shown in Equation (\ref{eq:tmp}). For $\mathrm{pr}^{(4)} \mathbf{v}_2$, we calculate its coefficients based on Equation (\ref{eq:prolong3}):
\begin{equation}
	\begin{cases}
		\phi^t = \mathrm{D}_t (-e^{-\frac{t}{t_0}} u_t) + e^{-\frac{t}{t_0}} u_{tt} = \frac{u_t}{t_0} e^{-\frac{t}{t_0}}, \\
		\phi^x = \mathrm{D}_x (-e^{-\frac{t}{t_0}} u_t) + e^{-\frac{t}{t_0}} u_{tx} = 0, \\ 
		\phi^{xx} = \mathrm{D}_{xx} (-e^{-\frac{t}{t_0}} u_t) + e^{-\frac{t}{t_0}} u_{txx} = 0, \\
		\phi^{xxx} = \mathrm{D}_{xxx} (-e^{-\frac{t}{t_0}} u_t) + e^{-\frac{t}{t_0}} u_{txxx} = 0, \\ 
		\phi^{xxxx} = \mathrm{D}_{xxxx} (-e^{-\frac{t}{t_0}} u_t) + e^{-\frac{t}{t_0}} u_{txxxx} = 0.
	\end{cases}
\end{equation}
Then, we have:
\begin{equation}
	\mathrm{pr}^{(4)} \mathbf{v}_2 = \mathbf{v}_2 + \frac{u_t}{t_0} e^{-\frac{t}{t_0}} \frac{\partial}{\partial u_t} = e^{-\frac{t}{t_0}} \frac{\partial}{\partial t} + \frac{u_t}{t_0} e^{-\frac{t}{t_0}} \frac{\partial}{\partial u_t}.
\end{equation}
For $\mathrm{pr}^{(4)} \mathbf{v}_3$, its coefficients are:
\begin{equation}
	\begin{cases}
		\phi^t = \mathrm{D}_t [1 - t_0 (e^\frac{t}{t_0} - 1 ) u_x ] + t_0 (e^\frac{t}{t_0} - 1) u_{tx} = - u_x e^\frac{t}{t_0}, \\
		\phi^x = \mathrm{D}_x [1 - t_0 (e^\frac{t}{t_0} - 1 ) u_x ] + t_0 (e^\frac{t}{t_0} - 1) u_{xx} = 0, \\ 
		\phi^{xx} = \mathrm{D}_{xx} [1 - t_0 (e^\frac{t}{t_0} - 1 ) u_x ] + t_0 (e^\frac{t}{t_0} - 1) u_{xxx} = 0, \\
		\phi^{xxx} = \mathrm{D}_{xxx} [1 - t_0 (e^\frac{t}{t_0} - 1 ) u_x ] + t_0 (e^\frac{t}{t_0} - 1) u_{xxxx} = 0, \\ 
		\phi^{xxxx} = \mathrm{D}_{xxxx} [1 - t_0 (e^\frac{t}{t_0} - 1 ) u_x ] + t_0 (e^\frac{t}{t_0} - 1) u_{xxxxx} = 0.
	\end{cases}
\end{equation}
Then, we get:
\begin{equation}
	\mathrm{pr}^{(4)} \mathbf{v}_3 = \mathbf{v}_3 - u_x e^\frac{t}{t_0} \frac{\partial}{\partial u_t} = t_0 (e^\frac{t}{t_0} - 1) \frac{\partial}{\partial x} + \frac{\partial}{\partial u} - u_x e^\frac{t}{t_0} \frac{\partial}{\partial u_t}.
\end{equation}
Similarly to Equation (\ref{eq:tmp2}), we exclude the variable $x$ based on $\mathrm{pr}^{(4)} \mathbf{v}_1$ and update the search space as $\{t, u, u_t, u_x, u_{xx}, u_{xxx}, u_{xxxx}\}$.

Construct the characteristic equation as shown in Equation (\ref{eq:di2}) based on $\mathrm{pr}^{(4)} \mathbf{v}_2$:
\begin{equation}
	e^\frac{t}{t_0} \mathrm{d} t = \frac{t_0}{u_t} e^\frac{t}{t_0} \mathrm{d} u_t.
\end{equation}
Integrating it yields the general solution:
\begin{equation}
	e^{-\frac{t}{t_0}} u_t = c.
\end{equation}
By replacing the terms $t$ and $u_t$ with integral constants, we update the search space as $\{e^{-\frac{t}{t_0}} u_t, u, u_x, u_{xx}, u_{xxx}, u_{xxxx}\}$.

For $\mathrm{pr}^{(4)} \mathbf{v}_3$, we construct the characteristic equation as:
\begin{equation}
	\mathrm{d} u = - \frac{1}{u_x} e^{-\frac{t}{t_0}} \mathrm{d} u_t.
\end{equation}
Integral yields:
\begin{equation}
	e^{-\frac{t}{t_0}} u_t + u u_x = c.
\end{equation}
Introducing it into the search space to replace $e^{-\frac{t}{t_0}} u_t$ and $u$, we obtain the final differential invariants as $\{e^{-\frac{t}{t_0}} u_t + u u_x, u_x, u_{xx}, u_{xxx}, u_{xxxx}\}$.

\section{Data generation}
\label{sec:data}

We select trajectory samples generated from $4$ initial conditions in the training dataset for equation discovery and use the L-BFGS optimizer with a learning rate of $0.1$ for training. During sparse regression, parameters smaller than the threshold are masked to $0$ upon convergence, and the optimizer is reset. For the KdV, KS, and nKdV equations, we set the threshold to $0.5$, while for Burgers equation, we set it to $5 \times 10^{-3}$. All methods share the above experimental settings to ensure a fair comparison.

\section{Implementation detail}
\label{sec:implementation}

We select trajectory samples generated from $4$ initial conditions in the training dataset for equation discovery and use the L-BFGS optimizer with a learning rate of $0.1$ for training. During sparse regression, parameters smaller than the threshold are masked to $0$ upon convergence, and the optimizer is reset. For the KdV, KS, and nKdV equations, we set the threshold to $0.5$, while for Burgers equation, we set it to $5 \times 10^{-3}$. All methods share the above experimental settings to ensure a fair comparison. We perform experiments on a single-core NVIDIA GeForce RTX 3090 GPU with available memory of $24,576$ MiB.


\end{document}